\documentclass[12pt]{article}
\pdfoutput=1

\usepackage[final,nonatbib]{nips_2017_mod}

\usepackage[utf8]{inputenc} % allow utf-8 input
\usepackage[T1]{fontenc}    % use 8-bit T1 fonts
\usepackage{hyperref}       % hyperlinks
\usepackage{url}            % simple URL typesetting
\usepackage{booktabs}       % professional-quality tables
\usepackage{amsfonts}       % blackboard math symbols
\usepackage{nicefrac}       % compact symbols for 1/2, etc.
\usepackage{microtype}      % microtypography
\usepackage{listings}		% long verbatims

\usepackage{amsmath}

\DeclareMathOperator*{\argmin}{arg\,min}
\usepackage{bm}

\usepackage{algorithm}
\usepackage{algorithmic}
\renewcommand{\algorithmiccomment}[1]{\hfill$\triangleright$ #1}

\usepackage{graphicx} % more modern
\usepackage{caption}
\usepackage{subcaption}
\usepackage{wrapfig}

\title{Gaussian Prototypical Networks for Few-Shot Learning on Omniglot
	}

\author{
  Stanislav Fort  \\ Stanford University
}

\begin{document}

\maketitle

\begin{abstract} 
We propose a novel architecture for $k$-shot classification on the Omniglot dataset. Building on prototypical networks, we extend their architecture to what we call \emph{Gaussian prototypical networks}. Prototypical networks learn a map between images and embedding vectors, and use their clustering for classification. In our model, a part of the encoder output is interpreted as a confidence region estimate about the embedding point, and expressed as a Gaussian covariance matrix. Our network then constructs a direction and class dependent distance metric on the embedding space, using uncertainties of individual data points as weights. We show that Gaussian prototypical networks are a preferred architecture over vanilla prototypical networks with an equivalent number of parameters. We report state-of-the-art performance in 1-shot and 5-shot classification both in 5-way and 20-way regime (for 5-shot 5-way, we are comparable to previous state-of-the-art) on the Omniglot dataset. We explore artificially down-sampling a fraction of images in the training set, which improves our performance even further. We therefore hypothesize that Gaussian prototypical networks might perform better in less homogeneous, noisier datasets, which are commonplace in real world applications.
\end{abstract} 

\section{Introduction}

\subsection{Few-shot learning}
Humans are able to learn to recognize new object categories on a single or small number of examples. This has been demonstrated in a wide range of activities from hand-written character recognition~\cite{Lake2011}, and motor control~\cite{motor}, to acquisition of high level concepts~\cite{omniglot}. Replicating this kind of behavior in machines is the motivation for studying few-shot learning.

Parametric deep learning has been performing well in settings with abundance of data. In general, deep learning models have a very high functional expressivity and capacity, and rely on being slowly, iteratively trained in a supervised regime. An influence of a particular example within the training set is therefore small, as the training is designed to capture the general structure of the dataset. This prevents rapid introduction of new classes after training.~\cite{deepmind}  

In contrast, few-shot learning requires very fast adaptation to new data. In particular, $k$-shot classification refers to a regime where classes unseen during training must be learned using $k$ labeled examples. Non-parametric models, such as k-nearest-neighbors (kNN) do not overfit, however, their performance strongly depends on the choice of distance metric.~\cite{Atkeson1997} Architectures combining parametric and non-parametric models, as well as matching training and test conditions, have been performing well on $k$-shot classification recently.

\subsection{Gaussian prototypical networks}

In this paper we develop a novel architecture based on prototypical networks used in \cite{prototype2017}, and train it and test it on the Omniglot dataset~\cite{omniglot}. Vanilla prototypical networks map images into embedding vectors, and use their clustering for classification. They divide a batch into \textit{support}, and \textit{query} images, and use the embedding vectors of the support set to define a class prototype -- a typical embedding vector for a given class. Proximity to these is then used for classification.

Our model, which we call the \textit{Gaussian prototypical network}, maps an image into an embedding vector, and an estimate of the image quality. Together with the embedding vector, a confidence region around it is predicted, characterized by a Gaussian covariance matrix. Gaussian prototypical networks learn to construct a direction and class dependent distance metric on the embedding space. We show that our model is a preferred way of using additional trainable parameters compared to vanilla prototypical networks.

Our goal is to show that by allowing our model to express its confidence in individual data points, we reach better results. We also experiment with intentionally corrupting parts of our dataset in order to explore the extendability of our method to noisy, inhomogeneous real world datasets, where weighting individual data points might be crucial for performance.

We report, to our knowledge, state-of-the-art performance in 1-shot and 5-shot classification both in 5-way and 20-way regime (for 5-shot 5-way, we are comparable to previous state-of-the-art) on the Omniglot dataset.~\cite{omniglot} By studying the response of our model to down-sampled data, we hypothesize that its advantage might be even higher in lower quality, inhomogeneous datasets.

This paper is structured as follows: We describe related work in Section~\ref{sec:related_work}. We then proceed to introduce our methods in Section~\ref{sec:methods}. The episodic training scheme is also presented there. We discuss the Omniglot dataset in Section~\ref{sec:datasets}, and our experiments in Section~\ref{sec:experiments}. Finally, our conclusions are presented in Section~\ref{sec:conclusion}.

\section{Related work}
\label{sec:related_work}
Non-parametric models, such as $k$-nearest neighbors (kNN), are ideal candidates for few-shot classifiers, as they allow for incorporation of previously unseen classes. However, they are very sensitive to the choice of distance metric.~\cite{Atkeson1997} Using the distance in the space of inputs directly (e.g. raw pixel values) does not produce high accuracies, as the connection between the image class and its pixels is very non-linear.

A straightforward modification in which a metric embedding is learned and then used for kNN classification has yielded good results, as demonstrated by \cite{metric1}, \cite{weinberger09distance}, \cite{metric2}, and \cite{metric3}. An approach using \textit{matching networks} has been proposed in \cite{matching}, in effect learning a distance metric between pairs of images. A noteworthy feature of the method is its training scheme, where each mini-batch (called an \textit{episode}) tries to mimic the data-poor test conditions by sub-sampling the number of classes as well as numbers of examples in each. It has been demonstrated that such an approach improves performance on few-shot classification.~\cite{matching} We therefore use it as well.

Instead of learning on the dataset directly, it has recently been proposed \cite{meta} to train an LSTM~\cite{LSTM} to predict updates to a few-shot classifier given an episode as its input. This approach is referred to as \textit{meta-learning}. Meta-learning has been reaching high accuracies on Omniglot~\cite{omniglot}, as demonstrated by \cite{fin}, and \cite{munk}. A task-agnostic meta-learner based on temporal convolutions has been proposed in~\cite{TCML}. Combinations of parametric and non-parametric methods have been the most successful in few-shot learning recently.~\cite{prototype2017}\cite{cdkNN}\cite{siam2015}

Our approach is specific to classification of images, and does not attempt to solve the problem via meta-learning. We build on the model presented in \cite{prototype2017}, which maps images into embedding vectors, and uses their clustering for classification. The novel feature of our model is that it predicts its confidence about individual data points via a learned, image-dependent covariance matrix. This allows it to construct a richer embedding space to which images are projected. Their clustering under a direction and class-dependent distance metric is then used for classification.

\section{Methods}
\label{sec:methods}
In this paper, we first explore the \textit{prototypical networks} described in \cite{prototype2017}. We extend the architecture to what we call a \textit{Gaussian prototypical network}, allowing the model to reflect quality of individual data points (images) by predicting their embedding vectors as well as confidence regions around them, characterized by a Gaussian covariance matrix.

A vanilla prototypical network comprises an encoder that maps an image into an embedding vector. A batch contains a subset of available training classes. In each iteration, images for each class are randomly split into \textit{support}, and \textit{query} images. The embeddings of support images are used to define class \textit{prototypes} -- embedding vectors typical of the class. The proximity of query image embeddings to the class prototypes is used for classification.

The encoder architectures of vanilla and Gaussian prototypical networks do not differ. The key difference is the way encoder outputs are interpreted, used, and how the metric on the embedding space is constructed. In Gaussian networks, a part of the encoder output is used to construct covariance matrices about embedding vectors, which allows our model to reflect the predictive power, and quality of individual data points.

\subsection{Encoder}
We use a multi-layer convolutional neural network without an explicit, final fully connected layer to encode images into high-dimensional Euclidean vectors. For a vanilla prototypical network described in \cite{prototype2017}, the encoder is a function taking an image $I$ and transforming it into a vector $\vec{x}$ as
\begin{equation}
\mathrm{encoder} (W): I \in \mathbb{R}^{H \times W \times C} \to \vec{x} \in \mathbb{R}^{D} \, ,
\end{equation}
where $H$ and $W$ are the height and width of the input image, and $C$ is the number of its channels. $D$ is the embedding dimension of our vector space which is a hyperparameter of the model. $W$ are the trainable weights of the encoder.

For a Gaussian prototypical network, the output of the encoder is a concatenation of an embedding vector $\vec{x} \in \mathbb{R}^{D}$ and the relevant components of the covariance matrix $\Sigma \in \mathbb{R}^{D \times D}$. Therefore
\begin{equation}
\mathrm{encoder}_{\mathrm{Gauss}} (W): I \in \mathbb{R}^{H \times W \times C} \to \left [ \vec{x}, \vec{s}_{\mathrm{raw}} \right ] \in \left [ \mathbb{R}^{D}, \mathbb{R}^{D_S} \right ] \, ,
\end{equation}
where $D_S$ is the dimensionality of the predicted components of the covariance matrix.

We explore three variants of the Gaussian prototypical network:
\begin{itemize}
	\label{item:type}
	\item[a)] \textbf{Radius} covariance estimate. $D_S = 1$ and only a single real number $s_\mathrm{raw} \in \mathbb{R}^1$ is generated per image to characterize the size of the confidence interval around its embedding vector. As such the covariance matrix has the form $\Sigma = \mathrm{diag}\left( \sigma,\sigma, \dots, \sigma \right)$, where $\sigma$ is calculated from the raw encoder output $s_\mathrm{raw}$. The confidence estimate is therefore not direction-sensitive. This method proved to be the most efficient usage of additional parameters on the Omniglot dataset \cite{omniglot}. We suspect that this preference might be dataset-specific, and that less homogeneous datasets could prefer more complex covariance estimates. 
	
	\item[b)] \textbf{Diagonal} covariance estimate. $D_S = D$ and the dimension of the covariance estimate is the same as of the embedding space. $\vec{s}_\mathrm{raw} \in \mathbb{R}^D$ is generated per image to characterize the size of the confidence interval around the embedding vector. Therefore the covariance matrix has the form $\Sigma = \mathrm{diag}\left( \vec{\sigma} \right)$, where $\vec{\sigma}$ is calculated from the raw encoder output $\vec{s}_\mathrm{raw}$. This allows the network to express direction-dependent confidence about a data point, although the confidence ellipsoid always remains axis-aligned with the embedding space axes.
	
	\item[c)] \textbf{Full} covariance estimate. A full covariance matrix is output per data point. This method proved to be needlessly complex for the tasks given and therefore was not explored further. 
\end{itemize} 

We used down-sampled gray-scale Omniglot images of the dimension $28 \times 28 \times 1$ as an input. A 4-layer CNN architecture with $2\times2$ max pooling results into a volume of shape $1 \times 1 \times (D + D_S)$, where the embedding dimension $D$ plus the relevant parts of the covariance matrix $D_S$ are equal to the number of filters in the last later. We were using the \verb|TensorFlow| \textit{SAME} padding and stride 1. Our filters were $3\times3$ in spatial extent. The final layer was equivalent to a fully-connected layer.

We were using 2 encoder architectures: 1) a \textbf{small} architecture, and 2) a \textbf{big} architecture. The small architecture corresponded to the one used in \cite{prototype2017}, and we used it to validate our own experiments with respect to the previous state-of-the-art results. The big architecture was used to see the effect of an increased model capacity on accuracy. As a basic building block, we used the sequence of layers in Equation~\ref{eq:block}.
\begin{equation}
\label{eq:block}
3\times3 \, \mathrm{CNN} \to \mathrm{batch\,normalization} \to \mathrm{ReLU} \to 2\times2 \, \mathrm{max\,pool} 
\end{equation}
Both architectures were composed of 4 such blocks stacked together. The details of the architectures are as follows:
\begin{itemize}
	\item[1)] \textbf{Small architecture:} $3 \times 3$ filters, numbers of filters $[64,64,64,D]$ ($[64,64,64,D+1]$ for the radius Gaussian model, $[64,64,64,2D]$ for the diagonal Gaussian model). Embedding space dimensions explored were $D = 32,64,128$.
	\item[2)] \textbf{Big architecture:} $3 \times 3$ filters, numbers of filters $[128,256,512,D]$ ($[128,256,512,D+1]$ for the radius Gaussian model, $[128,256,512,2D]$ for the diagonal Gaussian model). Embedding space dimensions explored were $D = 128,256,512$.
\end{itemize}

We explored 4 different methods of translating the raw covariance matrix output of the encoder into an actual covariance matrix. Since we primarily deal with the inverse of the covariance matrix $S = \Sigma^{-1}$, we were predicting it directly. Let the relevant part of the raw encoder output be $S_\mathrm{raw}$. The methods are as follows:  
\begin{itemize}
\label{item:sigma}	
	\item[a)] $S = 1 + \mathrm{softplus} \left ( S_\mathrm{raw} \right )$, where $\mathrm{softplus}(x) = \log \left ( 1 + e^x \right )$ and it is applied component-wise. Since $\mathrm{softplus}(x) > 0$, this guarantees $S > 1$ and the encoder can only make data points less important. The value of $S$ is also not limited from above. Both of these approaches prove beneficial for training. Our best models used this regime for initial training.
	\item[b)] $S = 1 + \mathrm{sigmoid} \left ( S_\mathrm{raw} \right )$, where $\mathrm{sigmoid}(x) = 1 / \left ( 1 + e^{-x} \right )$ and it is applied component-wise. Since $\mathrm{sigmoid}(x) > 0$, this guarantees $S > 1$ and the encoder can only make data points less important. The value of $S$ is bounded from above, as $S < 2$, and the encoder is therefore more constrained.
	\item[c)] $S = 1 + 4 \, \mathrm{sigmoid} \left ( S_\mathrm{raw} \right )$ and therefore $1 < S < 5$. We used this to explore the effect of the size of the domain of covariance estimates on performance.
	\item[d)] $S = \mathrm{offset} + \mathrm{scale} \times \mathrm{softplus} \left ( S_\mathrm{raw} / \mathrm{div} \right )$, where  $\mathrm{offset}$, $\mathrm{scale}$, and  $\mathrm{div}$ are initialized to $1.0$ and are trainable. Our best models used this regime for late-stage training, as it is more flexible and data-driven than a).  
\end{itemize}

\subsection{Episodic training}
\begin{figure}[t]
	\begin{center}
		\includegraphics[width=0.85\linewidth]{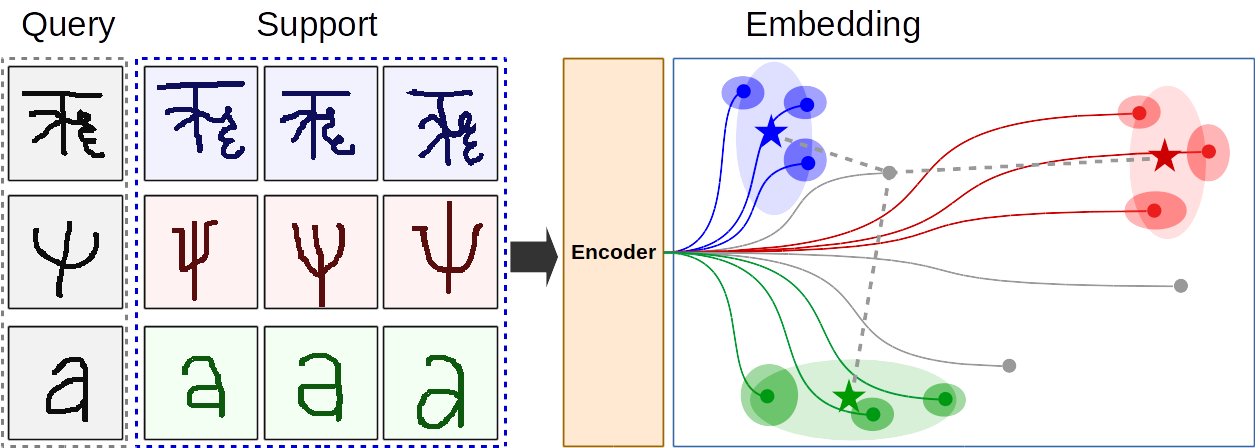}
	\end{center}
	\caption{A diagram of the function of the Gaussian prototypical network. An encoder maps an image into a vector in the embedding space (dark circles). A covariance matrix is also output for each image (dark ellipses). Support images are used to define the prototypes (stars), and covariance matrices (light ellipses) of the particular class. The distances between centroids, and encoded query images, modified by the total covariance of a class, are used to classify query images. The distances are shown as dashed gray lines for a particular query point.}
	\label{fig:diagram1}
\end{figure}

A key component of the prototypical model is the episodic training regime described in \cite{prototype2017}. During training, a subset of $N_c$ classes is chosen from the total number of classes in the training set (without replacement). For each of these classes, $N_s$ \textit{support} examples are chosen at random, as well as $N_q$ \textit{query} examples. The encoded embeddings of the support examples are used to define where a particular class \textit{prototype} lies in the embedding space. The distances between the query examples and positions of class prototypes are used to classify the query examples and to calculate loss. For the Gaussian prototypical network, the covariance of each embedding point is estimated as well. A diagram of the process is shown in Figure~\ref{fig:diagram1}. 

For a Gaussian prototypical network, the radius or a diagonal of a covariance matrix is output together with the embedding vector (more precisely its raw form is, as detailed in Section~\ref{item:sigma}). These are then used to weight the embedding vectors corresponding to support points of a particular class, as well as to calculate a total covariance matrix for the class. The distance $d_c(i)$ from a class prototype $c$ to a query point $i$ is then calculated as 
\begin{equation}
d_c^2(i) = \left ( \vec{x}_i - \vec{p}_c \right )^T S_c \left ( \vec{x}_i - \vec{p}_c \right ) \, ,
\end{equation}
where $\vec{p_c}$ is the centroid, or \textit{prototype}, of the class $c$, and $S_c = \Sigma_c^{-1}$ is the inverse of its covariance matrix. The Gaussian prototypical network is therefore able to learn class and direction-dependent distance metric in the embedding space. We found that the speed of training and its accuracy depend strongly on how distances are used to construct a loss. We conclude that the best option is to work with linear Euclidean distances, i.e. $d_c(i)$. The specific form of the loss function used is presented in Algorithm~\ref{alg:prototrain}. A diagram of the embedding space for a Gaussian prototypical network is shown in Figure~\ref{fig:diagram2}. A sample of the embedding space during training is shown in the Appendix in Figures~\ref{fig:embed} and~\ref{fig:embed_dots}. It illustrates the clustering of similar characters that is used for classification.
\begin{figure}[t]
	\begin{center}
		\includegraphics[width=0.45\linewidth]{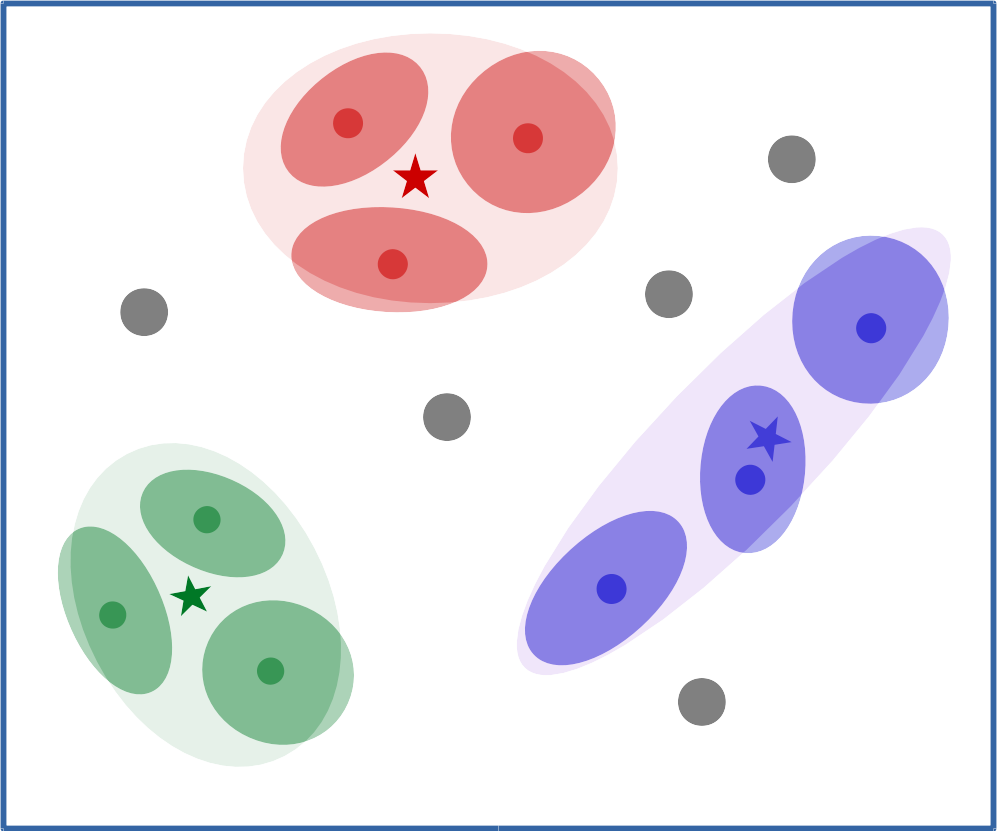}
	\end{center}
	\caption{A diagram showing the embedding space of a Gaussian prototypical network. An image is mapped to its embedding vector (dark dot) by the encoder. Its covariance matrix (dark ellipse) is also output by the encoder. An overall covariance matrix for each class is then computed (large light ellipses), as well as prototypes of the classes (stars). The covariance matrix of a class is used to locally modify the distance metric to query points (shown in gray).}
	\label{fig:diagram2}
\end{figure}

We study settings in which the covariance matrix is diagonal, as summarized in Section~\ref{item:type}. For the \textit{radius} case, $S = sI$, where $I$ is the identity matrix, and $s \in \mathbb{R}^1$ is calculated from the raw encoder output for each image. For the \textit{diagonal} case, $S = \mathrm{diag} \left ( \vec{s} \right )$, where $\vec{s}$ is similarly calculated from the raw encoder output for each image. 

\subsection{Defining a class}
A critical part of a prototypical network is the creation of a class prototype from the available support points of a particular class. We propose a variance-weighted linear combination of embedding vectors of individual support examples as our solution. Let class $c$ have support images $I_i$ that are encoded into embedding vectors $\vec{x}^c_i$, and inverses of covariance matrices $S^c_i$, whose diagonals are $\vec{s}^c_i$. The prototype, i.e. the centroid of the class, is defined as
\begin{equation}
\label{eq:prot}
\vec{p}_c = \frac{\sum_i \vec{s}^c_i \circ \vec{x}^c_i }{\sum_i \vec{s}^c_i} \, ,
\end{equation}  
where $\circ$ denotes a component-wise multiplication, and the division is also component-wise. The diagonal of the inverse of the class covariance matrix is then calculated as
\begin{equation}
\label{eq:sigma}
\vec{s}_c = \sum_i \vec{s}^c_i \, .
\end{equation}  
This corresponds to the optimal combination of Gaussians centered on the individual points into an overall class Gaussian, hence the name of the network. The elements of $s$ are effectively $1/\sigma^2$. Equations~\ref{eq:prot} and~\ref{eq:sigma} therefore correspond to weighting examples by $1/\sigma^2$. This allows the network to down-weight examples that are less important for defining the class, and therefore makes our architecture more suitable for noisy, inhomogeneous, or otherwise imperfect datasets.

For a one-shot regime, which is the way our networks were trained, there is a single labeled vector $\vec{x}^c$ defining each class. That means that the vector itself becomes the class prototype, and its covariance matrix is also inherited by the class. The covariance then comes into play in modifying distances to query points. The full algorithm is described in Algorithm~\ref{alg:prototrain}.

\begin{algorithm*}[tb]
	\caption{Classification and loss algorithm for Gaussian prototypical networks}
	\label{alg:prototrain}
	\begin{algorithmic}		
		\REQUIRE Images $I$, class labels $y$, encoder $f$, $N_s$ number of support points per class, $N_q$ number of query points per class, $N_c$ number of classes in a batch
		\ENSURE Predicted labels $\hat{y}$, loss $L$
		\FOR{batch in data}
			\STATE Choose a subset $C_\mathrm{used}$ of $N_c$ classes from all possible training classes at random (without replacement)
			\FOR{class $c$ in classes $C_\mathrm{used}$}
				\STATE Choose $N_s$ \textit{support} examples for the class $c$ from batch and call them $S_c$.
				\STATE Choose $N_q$ \textit{query} examples for the class $c$ from batch and call them $Q_c$.
				\STATE Embed $Q_c$ as $f(Q_c) \to \tilde{Q}^c, \_$ 
				\STATE Embed $S_c$ as $f(S_c) \to \tilde{S}^c , \{ \vec{s}^c \}$ \COMMENT{vectors and covariances}
				\STATE Class prototype $\vec{p}_c \gets \left ( \sum_i \vec{s}^c_i \circ \tilde{Q}^c_i \right ) / \left ( \sum_i \vec{s}^c_i \right ) $ \COMMENT{summed over query points}
				\STATE Class inverse covariance $\vec{s}^c = \sum_i \vec{s}^c_i$ \COMMENT{summed over query points} 
			\ENDFOR 
			\STATE Loss $L \gets 0$ \COMMENT{zeroing the loss per batch}
			\FOR{query image $i$ in batch query points}
				\STATE Let the image $i$ embedding vector be $\vec{x}$ and its true class $y$
				\FOR{class $c$ in classes $C_\mathrm{used}$}
					\STATE $\mathrm{difference}(c,i) =  \vec{x} - \vec{p}_c$ \COMMENT{position difference from class $c$}
					\STATE $\mathrm{distance}(c,i) = \sqrt{ \left ( \vec{x} - \vec{p}_c \right )^T \vec{s}^c_i \circ \left ( \vec{x} - \vec{p}_c \right ) }$ \COMMENT{covariance-modified distance}
				\ENDFOR
				\STATE Predicted label $\hat{y}_i \gets \argmin_c \mathrm{distance}(c,i)$ \COMMENT{labelled based on the "closest" prototype}
				\STATE $L \gets L + \frac{1}{N_c} \mathrm{softmax\,cross\,entropy}(-\mathrm{distance}(c,i), y)$ 
				\COMMENT{summed over $c$}
			\ENDFOR
			\STATE $L \gets L / \mathrm{batch\,size}$ \COMMENT{loss per batch for comparability} 
		\ENDFOR
	\end{algorithmic}
\end{algorithm*}

\subsection{Evaluating models}
\label{sec:evaluate}
To estimate the accuracy of a model on the test set, we classify the whole test set for every number of support points $N_s = k$ in the range $k \in [1,..19]$. The number of query points for a particular $k$ is therefore $N_q = 20 - N_s$, as Omniglot provides 20 examples of each class. The accuracies are then aggregated, and for a particular stage of the model training a $k$-shot classification accuracy as a function of $k$ is determined. Since we are not using a designated validation set, we ensure our impartiality by considering the test results for the 5 highest training accuracies, and calculate their mean and standard deviation. By doing that, we prevent optimizing our result for the test set, and furthermore obtain error bounds on the resulting accuracies. We evaluate our models in 5-way and 20-way test classification to directly compare to existing literature.

\section{Datasets}
\label{sec:datasets}
We used the Omniglot dataset.~\cite{omniglot} Omniglot contains 1623 character classes from 50 alphabets (real and fictional) and 20 hand-written, gray-scale, $105 \times 105$ pixel examples of each. We down-sampled them to $28 \times 28 \times 1$, subtracted their mean, and inverted them. We were using the recommended split to 30 training alphabets, and 20 test alphabets, as suggested by \cite{omniglot}, and used by \cite{prototype2017}. The training set included overall 964 unique character classes, and the test set 659 of them. There was no class overlap between the training and test datasets. We did not use a separate validation set as we did not fine-tune hyperparameters and chose the best performing model based on training accuracies alone (see Section~\ref{sec:evaluate}).

\begin{figure}[t]
	\begin{center}
		\includegraphics[width=0.4\linewidth]{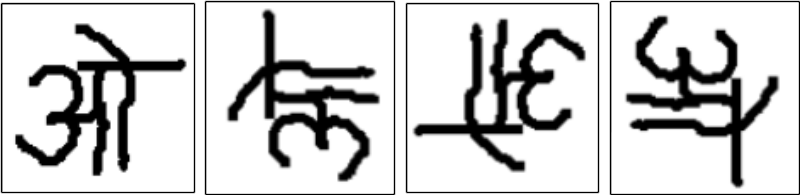}
	\end{center}
	\caption{An example of augmentation of class count by rotations. An original character (on the left) is rotated by $90^\circ$, $180^\circ$, and $270^\circ$. Each rotation is then defined as a new class. This enhances the number of classes, but also introduces degeneracies for symmetric characters.}
	\label{fig:rot}
\end{figure}

To extend the number of classes, we augmented the dataset by rotating each character by $90^\circ$, $180^\circ$, and $270^\circ$, and defined each rotation to be a new character class on its own. The same approach is used in \cite{matching}, and \cite{prototype2017}. An example of an augmented character is shown in Figure~\ref{fig:rot}. This increased the number of classes 4-fold. In total, the training set therefore included 77,120 images, and the test set 52,720 images. Due to the rotational augmentation, characters that have a rotational symmetry were nonetheless defined as multiple classes. As even a hypothetical perfect classifier would not be able to differentiate e.g. the character "O" from a rotated "O", $100\,\%$ accuracy was not reachable.

To improve training, and utilization of the ability to predict covariances of characters by the Gaussian network, we purposefully down-sampled a part of the training set in some of our experiments. Details are provided in Section~\ref{sec:experiments}. Our results suggest that the Omniglot dataset is too simple to fully utilize the ability of the Gaussian network to estimate covariance matrices. We hypothesize that the full strength of our method would show more on inhomogeneous datasets with varying quality of individual data points, as is commonly the case in real world applications. 

\section{Experiments}
\label{sec:experiments}
We conducted a large number of few-shot learning experiments on the Omniglot dataset. For Gaussian prototypical networks, we explored different embedding space dimensionalities, ways of generating the covariance matrix, and encoder capacities (see Section~\ref{item:sigma} for details). We also compared them to vanilla prototypical networks, and showed that our Gaussian variant is favorable, and that in particular the most efficient way of using the additional parameters is to predict a single number per embedding point (the radius method in Section~\ref{item:sigma}).  In general, we explored the size of the encoder (small, and big, as described in Section \ref{sec:methods}), the Gaussian/vanilla prototypical network comparison, the distance metric (cosine, $\sqrt{L_2}$, $L_2$, and $L_2^2$), the number of degrees of freedom of the covariance matrix in the Gaussian networks (radius, and diagonal estimates, see Section~\ref{item:sigma}), and the dimensionality of the embedding space. We also explored augmenting the input dataset by down-sampling a subset of it to encourage the usage of covariance estimates by the network, and found that this improves $(k>1)$-shot performance. 

We were using the \textit{Adam} optimizer with an initial learning rate of $2 \times 10^{-3}$. We halved the learning rate every $2000$ episodes $\approx$ $30$ epochs. All our models were implemented in \verb|TensorFlow|, and ran on a single NVidia K80 GPU on Google Cloud. The training time of each model was less than a day.

We trained our models with $N_c = 60$ classes (60-way classification) at training time, and tested on $N_{ct} = 20$ classes (20-way) classification. For our best-performing models, we also conducted a final $N_{ct} = 5$ (5-way) classification test to compare our results to literature. During training,  each class present in the mini-batch comprised $N_s = 1$ support points, as we found that limiting the number of support points leads to better classification accuracies. This could intuitively be understood as matching the training regime to the test regime. The remaining $N_q = 20 - N_s = 19$ images per class were used as query points.

The detailed results of our experiments are summarized in Table~\ref{tab:results1}. We explored 4 ways of estimating the covariance matrix from the raw covariance output of the encoder, as detailed in Section~\ref{item:type}.

\begin{figure}[t]
	\begin{center}
		\includegraphics[width=0.65\linewidth]{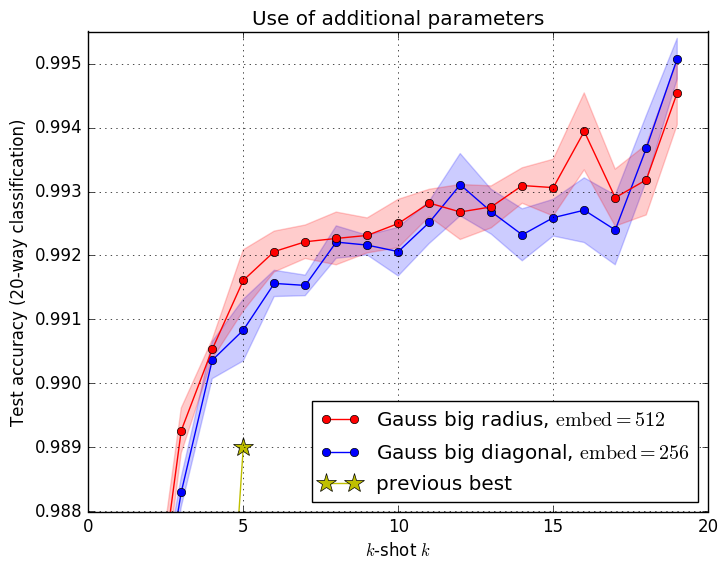}
	\end{center}
	\caption{Comparison of two methods of allocation of extra parameters. Allocating extra parameters to increase embedding space dimensionality (radius), or making a more precise covariance estimate (diagonal). The radius estimate (1 additional real number per embedding vector) outperforms the diagonal estimate, and the vanilla prototypical network with the same number of parameters.}
	\label{fig:type_effect}
\end{figure}

\begin{figure}[t]
	\begin{center}
		\includegraphics[width=0.65\linewidth]{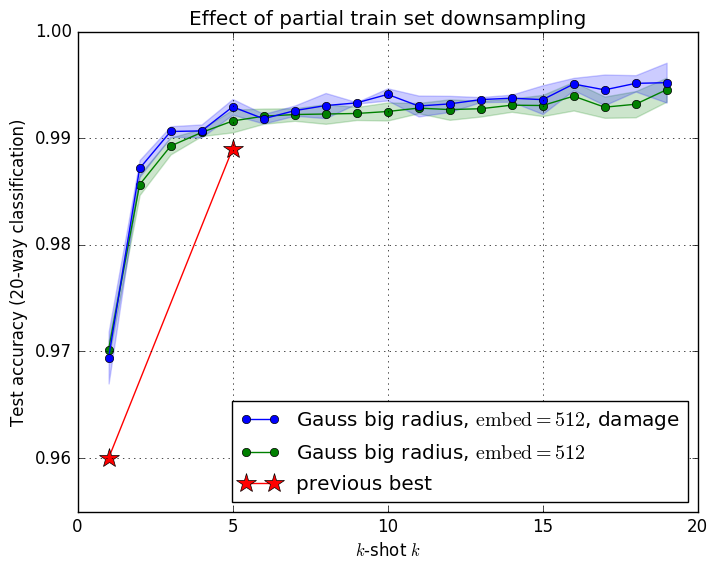}
	\end{center}
	\caption{The effect of down-sampling a part of the training set on $k$-shot test accuracy. The version trained on purposefully damaged data outperforms the one trained on unmodified data, as it learned to utilize covariance estimates better.}
	\label{fig:damage}
\end{figure}

\begin{figure}[t]
	\begin{center}
		\includegraphics[width=0.65\linewidth]{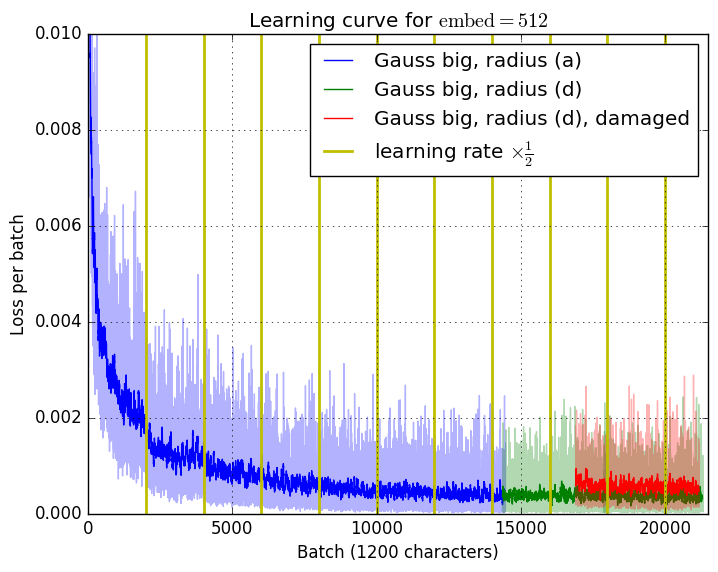}
	\end{center}
	\caption{The loss as a function of iteration. The yellow vertical lines show where the learning rate was halved. The beneficial effect of learning rate halving is visible at the beginning. The red segment corresponds to training on a partially down-sampled training set, and therefore has a higher loss.}
	\label{fig:lr}
\end{figure}

\begin{figure}[t]
	\begin{center}
		\includegraphics[width=0.65\linewidth]{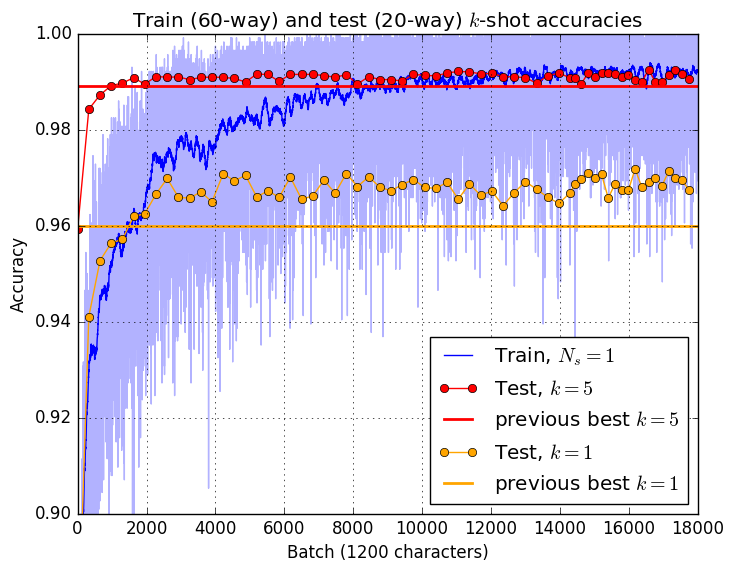}
	\end{center}
	\caption{Training accuracy compared to test accuracy. The plot shows the training accuracy (60-way classification) of a large Gaussian prototypical network (radius covariance estimate) and compares it to its 1-shot and 5-shot test performance (20-way classification). It also compares the results to the current state-of-the-art.~\cite{prototype2017}}
	\label{fig:accs}
\end{figure}

We also verified, provided that the covariance estimate is not needlessly complex, that using encoder outputs as covariance estimates is more advantageous than using the same number of parameters as additional embedding dimension. This holds true for the \textit{radius} estimate (i.e. one real number per embedding vector), however, the \textit{diagonal} estimate does not seem to help with performance (keeping the number of parameters equal). This effect is shown in Figure~\ref{fig:type_effect}, and Table~\ref{tab:results1}.
The best performing model was initially trained on the undamaged dataset for 220 epochs. The training then continued with $1.5\%$ of images down-sampled to $24 \times 24$, $1.0\%$ down-sampled to $20 \times 20$, and $0.5\%$ down-sampled to $16 \times 16$ for 100 epochs. Then with $1.5\%$ down-sampled to $23 \times 23$ and $1.0\%$ down-sampled to $17 \times 17$ for 20 epochs, and $1.0\%$ down-sampled to $23 \times 23$ for 10 epochs. These choices were quite arbitrary and not optimized over. The purposeful damage to the dataset encouraged usage of the covariance estimate and increased $(k>1)$-shot results, as shown in Table~\ref{tab:results1}, and Figure~\ref{fig:damage}. This partially shows that the Omniglot dataset is too high quality and simple a testbed for our approach. The training loss curves are shown in Figure~\ref{fig:lr}.
The training, and test accuracies as functions of iteration are also shown in Figure~\ref{fig:accs}.
\begin{table*}
	\begin{center}
		\centerline{
			\begin{tabular}{|l|l|c|c||c|c|}
				\hline
				\multicolumn{4}{|c||}{} & \multicolumn{2}{|c|}{20-way} \\
				\hline
				Model & Covariance & Damage & Embedding dim. & 1-shot & 5-shot \\
				\hline
				Vanilla & & & 512  &  0.9697 $\pm$ 0.0004 \%  &  0.9913 $\pm$ 0.0003 \% \\
				Big Gauss &  Diagonal (a) & & 256   &  0.9688 $\pm$ 0.0008 \% &  0.9910 $\pm$ 0.0004 \% \\
				Big Gauss &  Radius (b) & & 512  &  0.9703 $\pm$ 0.0003 \% &  0.9917 $\pm$ 0.0005 \% \\
				Big Gauss &  Radius (c) & & 512   &  0.9698 $\pm$ 0.0003 \%  &  0.9915 $\pm$ 0.0005 \% \\
				Big Gauss &  Radius (d) & & 512  &  \textbf{0.9702 $\pm$ 0.0004 \%}  &  \textbf{0.9916 $\pm$ 0.0005 \%}  \\
				Big Gauss &  Radius (d) & Yes & 512  &  \textbf{0.9694 $\pm$ 0.0014 \%}  &  \textbf{0.9929 $\pm$ 0.0004 \%}  \\
				\hline
				\multicolumn{4}{|c||}{} & \multicolumn{2}{|c|}{5-way} \\
				\hline
				Big Gauss &  Radius (d) & & 512  &  \textbf{0.9902 $\pm$ 0.0005 \%}  &  \textbf{0.9966 $\pm$ 0.0002 \%}  \\
				Big Gauss &  Radius (d) & Yes & 512  &  \textbf{0.9907 $\pm$ 0.0003 \%}  &  \textbf{0.9973 $\pm$ 0.0002 \%}  \\		
				\hline
			\end{tabular} 
		}
	\end{center}
	\caption{The test results for the big encoder architecture ($3 \times 3$ filters, 4 layers, number of filters = 128,256,512,$\bullet$) comparing the effect of dimensionality of the covariance matrix as well as the embedding space on the final accuracy. (a, b, c, d) relates to different methods of converting the raw encoder output to the covariance matrix. The radius estimate of the covariance adds one more dimension to the encoder output. The diagonal estimate doubles the number of encoder outputs. The \textit{Big Gauss} with embedding dimension $256$ and a diagonal covariance therefore has the same number of parameters as the vanilla network with $512$. The radius estimate adds 1 dimension and is therefore comparable to the vanilla model of the same embedding dimensionality. The damage column signifies that the training set was purposefully partially down-sampled during training.}
	\label{tab:results1}
\end{table*}

\begin{figure}[t]
	\begin{center}
		\includegraphics[width=0.65\linewidth]{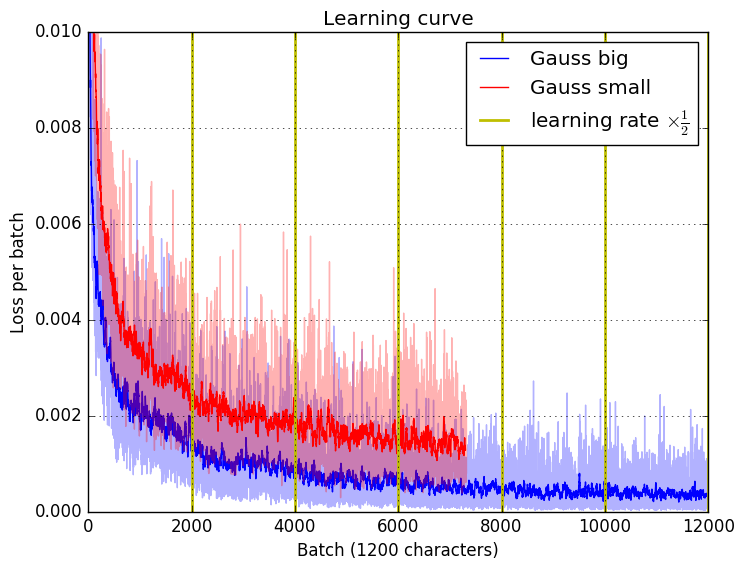}
	\end{center}
	\caption{The effect of model capacity on loss. The bigger model trains faster and reaches a smaller loss overall. The yellow vertical lines show where the learning rate was halved.}
	\label{fig:size}
\end{figure}

We conducted verification experiments with the small architecture, and reached comparable results to \cite{prototype2017}, as summarized in Table~\ref{tab:verification}. The table also shows that training in a $N_s > 1$ regime, i.e. with more data points defining a class, leads to worse performance. The effect of the higher capacity of the big model is shown in Figure~\ref{fig:size}.
\begin{table*}
	\begin{center}
		\begin{tabular}{|l||c|c|}
			\hline
			Method (20-way classification) & 1-shot test & 5-shot test \\
			\hline
			Best results from \cite{prototype2017} & $96.0\, \%$ & $98.9\, \%$ \\
			\hline
			Small, $N_s = 5$ & $94.28 \pm 0.26 \, \%$ & $98.61 \pm 0.18 \, \%$ \\ 
			Small, $N_s = 1$ & $95.64 \pm 0.22 \, \%$ & $98.87 \pm 0.11 \, \%$ \\ 
			Small Gauss $\sigma \in \mathbb{R}^1$, $N_s = 1$ & $95.99 \pm 0.28 \, \%$ & $98.81 \pm 0.16$ \% \\ 
			\hline
		\end{tabular}
	\end{center}
	\caption{Results of our verification experiments with small architectures. State of the art for 20-way classification 1-shot was $96.0 \, \%$ and for 5-shot $98.9 \, \%$. $N_s$ is the number of support points per class during training. All training was done in $N_c = 60$ (60-way classification) regime. For the Gaussian prototypical model $\sigma \in \mathbb{S}$ shows the dimensionality of the estimated covariance matrix.}
	\label{tab:verification}
\end{table*}
The comparison of our models, and results from literature is presented in Table~\ref{tab:comparison}. To our knowledge, our models outperform state-of-the-art 1-shot and 5-shot results in both 5-way and 20-way test-time classification on Omniglot. In 5-way 5-shot classification in particular, we are reaching very close to perfect performance ($99.73 \pm 0.02$ \%) and therefore conclude that a more complex dataset is needed for further few-shot learning algorithms development. 
\begin{table*}
	\begin{center}
		\centerline{
		\begin{tabular}{|l||c|c|c|c|}
			\hline
			 & \multicolumn{2}{c|}{{20-way}} & \multicolumn{2}{c|}{{5-way}}  \\
			\hline
			Model & 1-shot & 5-shot & 1-shot & 5-shot  \\
			\hline
			Matching networks \cite{matching} & 93.8\% & 98.5\% & 98.1\% & 98.9\% \\
			Matching networks \cite{matching} & $93.5\, \%$ & $98.7\, \%$ & 97.9\% & 98.7\% \\
			Neural statistician \cite{stats} & $93.2\, \%$ & $98.1\, \%$ & 98.1\% & 99.5\% \\
			Prototypical network \cite{prototype2017} & $96.0\, \%$ & $98.9\, \%$ & 98.8\% & \textbf{99.7}\% \\
			TCML \cite{TCML} & & & 98.8 $\pm$ 0.22\% & 99.2 $\pm$ 0.17\% \\ 
			Fin et al. \cite{fin} & & & 98.7 $\pm$ 0.4\% & \textbf{99.9 $\pm$ 0.3}\% \\
			Munkhdalai and Yu \cite{munk} & & & 98.9 \% & \\
			\textbf{Big Gauss (radius) (ours)} & \textbf{97.02 $\pm$ 0.04\%} & \textbf{99.16 $\pm$ 0.05\%} & \textbf{99.02 $\pm$ 0.05\%} & \textbf{99.66 $\pm$ 0.02\%}\\
			\textbf{Big Gauss (radius) with damage (ours)} & \textbf{96.94 $\pm$ 0.14\%} & \textbf{99.29 $\pm$ 0.04\%} & \textbf{99.07 $\pm$ 0.03\%} & \textbf{99.73 $\pm$ 0.02\%}\\
			\hline
		\end{tabular} 
}	
	\end{center}
	\caption{The best results of our experiments as compared to other papers. All training was done in $N_c = 60$ (60-way classification) regime. To our knowledge, our models have statistically significant state-of-the-art performance in both 1-shot and 5-shot 20-way classification, as well as for 1-shot 5-way classification. We perform comparably with current state-of-the-art in the 5-shot 5-way case.}
	\label{tab:comparison}
\end{table*}

\subsection{Usage of covariance estimate}
In order to validate our assumption that the Gaussian prototypical network outperforms the vanilla version due to its ability to predict covariances of individual embedded images and therefore the possibility to down-weight them, we studied the distribution of predicted values of $s$ for our best performing network that was trained on partially down-sampled training set. We purposefully down-sampled a part of the data, and studied the resulting distribution of covariances.

Down-sampling an image changes its mean and variance. As our encoders were built with a batch normalization layer in each block (see Equation~\ref{eq:block} for details), the meaning of a particular value of the raw output changes based on the current batch. Since our model was trained with batch normalization, turning it off to study the covariances would lead to irrelevant results. 

\begin{figure}[!htb]
	\begin{center}
		\includegraphics[width=0.55\linewidth]{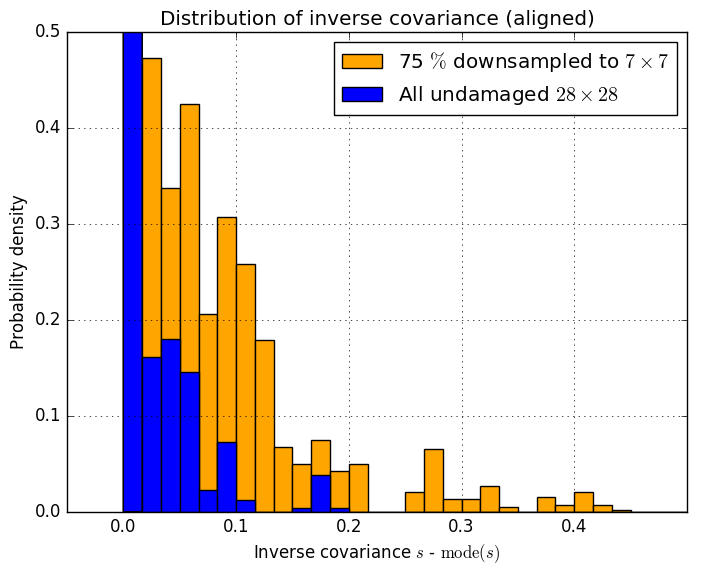}
	\end{center}
	\caption{Predicted covariances for the original dataset and a partially down-sampled version of it. The Gaussian network learned to down-weight damaged examples by predicting a higher $s$, as apparent from the heavier tail of the yellow distribution. The distributions are aligned together, as only the difference between the leading edge and a value influence classification.}
	\label{fig:sigma}
\end{figure}

For the undamaged dataset, the vast majority of covariance estimates took the same value. This stays true even when artificially introducing damage by down-sampling. However, the distributions are shifted due to the effect of batch normalization in the last layer. To better represent the meaning of individual inverse covariances, we aligned our histograms such that the most frequent values match each other. This approach is useful as the most dominant values correspond to the raw output of $0$, and only the differences from it influence classification. The result is shown in Figure~\ref{fig:sigma}.

\section{Conclusion}
\label{sec:conclusion}
In this paper we proposed Gaussian prototypical networks for few-shot classification -- an improved architecture based on prototypical networks~\cite{prototype2017}. We tested our models on the Omniglot dataset, and explored different approaches to generating a covariance matrix estimate together with an embedding vector. We showed that the Gaussian prototypical network outperforms the vanilla prototypical network with a comparable number of parameters, and therefore that our architecture choice is beneficial. We found that estimating a single real number on top of an embedding vector works better than estimating a diagonal or a full covariance matrix. We suspect that lower quality, less homogeneous datasets might prefer a more complex covariance matrix estimate. Contrary to \cite{prototype2017}, we found that the best results are obtained if one trains the network in the 1-shot regime. We then extended the size of our model and managed to reach, to the best of our knowledge, state-of-the-art performance in 1-shot and 5-shot classification both in 5-way and 20-way test regime (for 5-shot 5-way, we are comparable to previous state-of-the-art). We managed to get better accuracies (in particular for $(k>1)$-shot classification) by artificially down-sampling fractions of our training dataset, encouraging the network to fully utilize covariance estimates. Especially for 5-way classification, our results are very close to perfect performance and we therefore conclude that further development in few-shot classification should focus on more complex datasets than Omniglot. We hypothesize that the ability to learn the embedding as well as its uncertainty would be even more beneficial for poorer-quality datasets, which are commonplace in real world applications. There, down-weighting some data points might be crucial for faithful classification. This is supported by our experiments with down-sampling Omniglot.

\clearpage
\subsubsection*{Acknowledgements}
We would like to thank Ben Poole, and Yihui Quek (both at Stanford University) for useful discussions, and brainstorming. A part of this work was done as a class project for the Stanford University CS~231N: \textit{Convolutional Neural Networks for Visual Recognition}, which provided Google Cloud credit coupons that partially supported our GPU usage.  

{
	\small
	\bibliographystyle{unsrt-phys}
	\bibliography{gauss_lib}
	\bibliographystyle{unsrt-phys}
}

\newpage
\appendix

\section{Embedding space snapshots}

\begin{figure}[!htb]
	\begin{center}
		\includegraphics[width=0.7\linewidth]{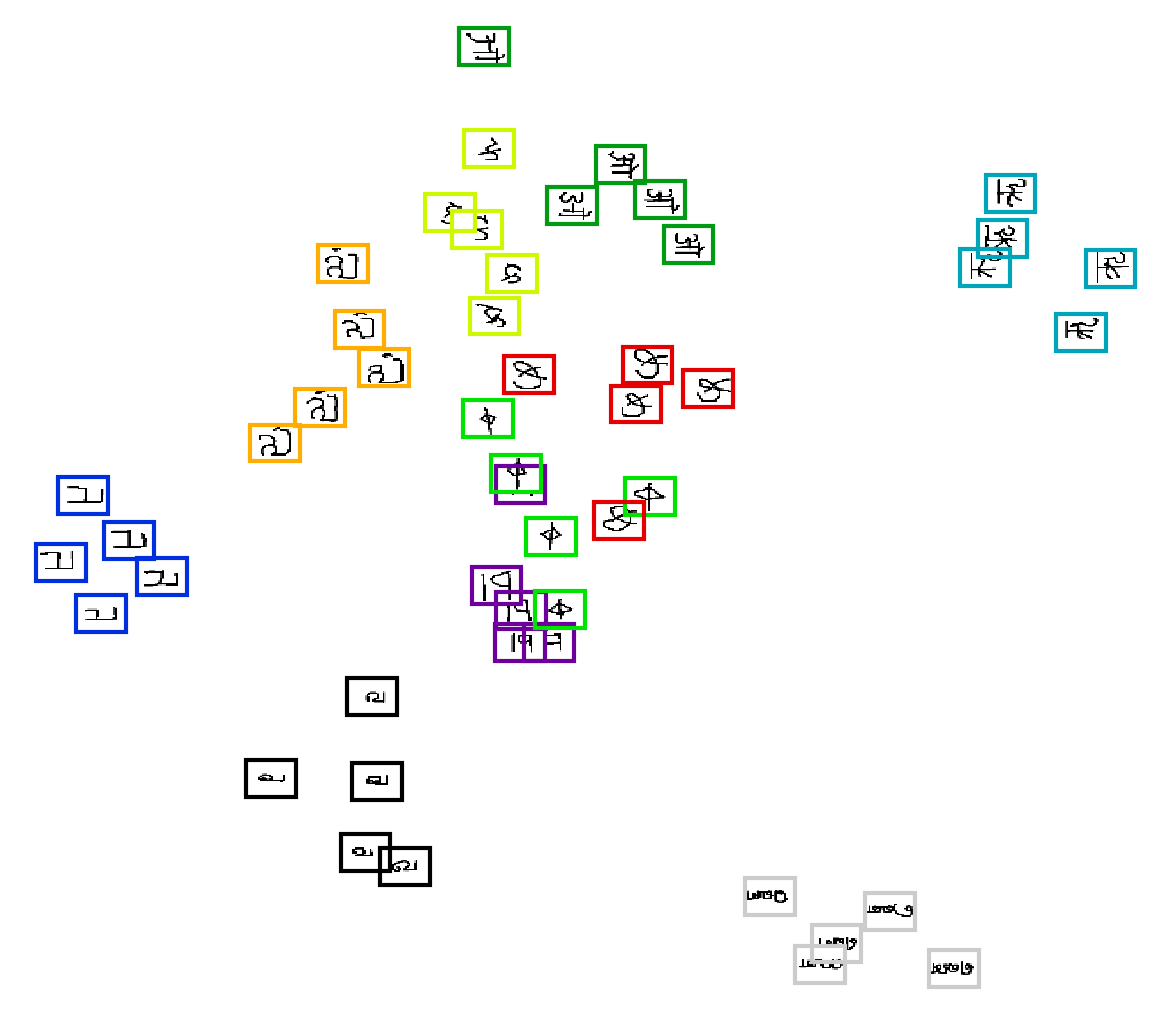}
	\end{center}
	\caption{A two-dimensional projection of the embedding space during training, as obtained by Principle Component Analysis. The clustering of similar characters, used for classification of unknown query images, is apparent in the plot.}
	\label{fig:embed}
\end{figure}

\begin{figure}[!htb]
	\begin{center}
		\includegraphics[width=0.7\linewidth]{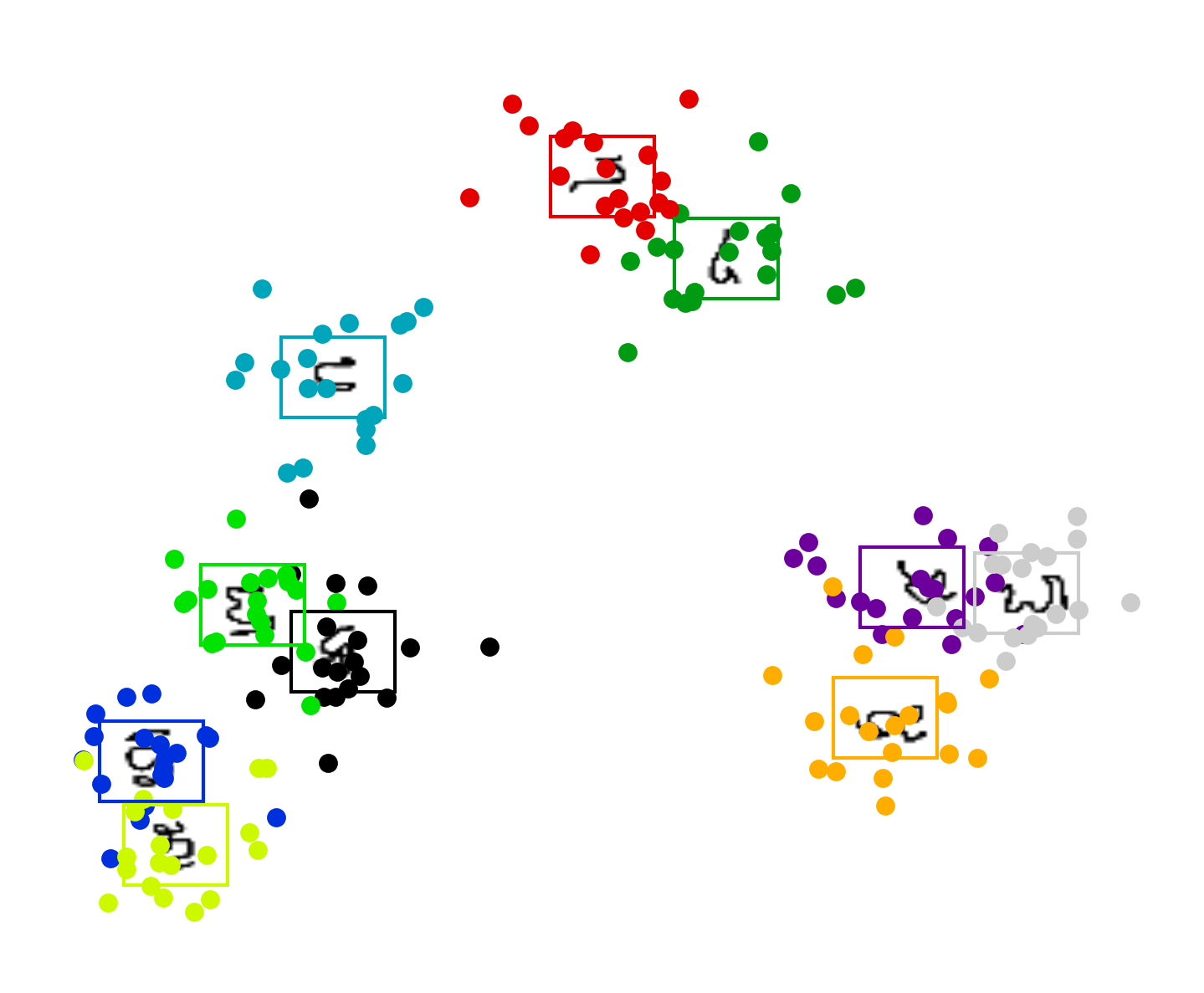}
	\end{center}
	\caption{A two-dimensional projection of the embedding space during training, as obtained by Principle Component Analysis. The dots represent embedding vectors of query points. Their clustering around class prototypes (characters) is apparent.}
	\label{fig:embed_dots}
\end{figure}

\end{document}